\documentclass[times,twoside]{SilvaLab_arxiv}

\usepackage{amsmath}
\usepackage{amssymb}
\usepackage{amsthm}
\usepackage{amsfonts}
\usepackage{verbatim}
\usepackage{authblk}
\usepackage{graphicx}

\usepackage{amsmath,amsfonts,bm}

\def\eqref#1{equation~\ref{#1}}

\def\1{\bm{1}}

\def\va{{\bm{a}}}

\def\vp{{\bm{p}}}

\def\vx{{\bm{x}}}

\def\mA{{\bm{A}}}

\def\mH{{\bm{H}}}

\def\mX{{\bm{X}}}

\DeclareMathAlphabet{\mathsfit}{\encodingdefault}{\sfdefault}{m}{sl}
\SetMathAlphabet{\mathsfit}{bold}{\encodingdefault}{\sfdefault}{bx}{n}

\def\gG{{\mathcal{G}}}

\leadauthor{Wang} 

\begin{document}

\title{Generalizable Machine Learning in Neuroscience using Graph Neural Networks}
\shorttitle{Generalizable Machine Learning using GNNs}

\author[1 \Letter]{Paul Y Wang}
\author[2,5]{Sandalika Sapra}
\author[3,5]{Vivek Kurien George}
\author[3,4,5 \Letter]{Gabriel A. Silva}

\affil[1]{Department of Physics, University of California San Diego}
\affil[2]{Department of Electrical and Computer Engineering, University of California San Diego}
\affil[3]{Department of Bioengineering, University of California San Diego}
\affil[4]{Department of Neuroscience, University of California San Diego}
\affil[5]{Center for Engineered Natural Intelligence, University of California San Diego}

\maketitle

\begin{abstract}
Although a number of studies have explored deep learning in neuroscience, the application of these algorithms to neural systems on a microscopic scale, i.e. parameters relevant to lower scales of organization, remains relatively novel. Motivated by advances in whole-brain imaging, we examined the performance of deep learning models on microscopic neural dynamics and resulting emergent behaviors using calcium imaging data from the nematode C. elegans. We show that neural networks perform remarkably well on both neuron-level dynamics prediction, and behavioral state classification. In addition, we compared the performance of structure agnostic neural networks and graph neural networks to investigate if graph structure can be exploited as a favourable inductive bias. To perform this experiment, we designed a graph neural network which explicitly infers relations between neurons from neural activity and leverages the inferred graph structure during computations. In our experiments, we found that graph neural networks generally outperformed structure agnostic models and excel in generalization on unseen organisms, implying a potential path to generalizable machine learning in neuroscience. 
\end {abstract}


\begin{corrauthor}
PW (pwang\at ucsd.edu) and GS (gsilva\at ucsd.edu)
\end{corrauthor}

\section*{Introduction}
Constructing generalizable models in neuroscience poses a significant challenge because systems in neuroscience are typically complex in the sense that dynamical systems composed of numerous components collectively participate to produce emergent behaviors. Analyzing these systems can be difficult because they tend to be highly non-linear in how they interact, can exhibit chaotic behaviors and are high-dimensional by definition. As such, indistinguishable macroscopic states can arise from numerous unique combinations of microscopic parameters, i.e. parameters relevant to lower scales of organization. Thus, bottom-up approaches to modeling neural systems often fail since a large number of microscopic configurations can lead to the same observables \cite{failure} \cite{similar}.

Because neural systems are highly degenerate and complex, their analysis is not amenable to many conventional algorithms. For example, observed correlations between individual neurons and behavioral states of an organism may not generalize to other organisms or even to repeated trials in the same individual \cite{fregnac2017big} \cite{churchland2010cortical} \cite{robustness-global-structure}. Hence, individual variability of neural dynamics remains poorly understood and a fundamental obstacle to model development, as evaluation on unseen individuals often leads to subpar results. Nevertheless, neural systems exhibit universal behavior: organisms behave similarly. Motivated by the need for robust and generalizable analytical techniques, researchers recently applied tools from dynamical systems analysis to simple organisms in hopes of discovering a universal organizational principle underlying behavior. These studies, made possible by advances in whole-brain imaging, reveal that neural dynamics live on low-dimensional manifolds which map to behavioral states \cite{sim-imaging} \cite{kato}. This discovery implies that although microscopic neural dynamics differ between organisms, a macroscopic/global universal framework may enable generalizable algorithms in neuroscience. Nevertheless, the need for significant hand-engineered feature extraction in these studies underscores the potential of deep learning models for scalable analysis of neural dynamics. 

In this work, we examine the performance and generalizability of deep learning models applied to the neural activity of C. elegans (round worm/nematode). In particular, C. elegans is a canonical species for investigating microscopic neural dynamics because it remains the only organism whose connectome (the mapping of all 302 neurons and their synaptic connections) is completely known and well studied \cite{white} \cite{bargmann2013connectome} \cite{varshney2011structural} \cite{connectome}. Furthermore, the transparent body of these worms allows for calcium imaging of whole brain neural activity which remains the only imaging technique capable of spatially resolving the dynamics of individual neurons \citep{wba}. Leveraging these characteristics and insight gained from previous studies, we developed deep learning models that bridge recent advances in neuroscience and deep learning. Specifically, we first demonstrate state-of-the-art performance for classifying motor action states of C. elegans from calcium imaging data acquired in previous works. Next, we examine the generalization performance of our deep learning models on unseen worms both within the same study and in worms from a separate study published years later. We then show that graph neural networks exhibit a favourable inductive bias for analyzing both higher-order function and microscopic/neuron-level dynamics in C. elegans.

\section*{Background}
In this section we discuss recent advances in neuroscience and machine learning upon which we build our model and experiments.

\subsection*{Universality/Generalizability in C. elegans models}
The motor action sequence of C. elegans is one of the only systems for which experiments on whole-brain microscopic neural activity may be performed and readily analyzed. As such, numerous efforts have focused on building models that can accurately capture the hierarchical nature of neural dynamics and resulting locomotive behaviors \cite{openworm} \cite{c302}. Taking advantage of this, Kato \emph{et. al.} \cite{kato} investigated neural dynamics corresponding to a pirouette, a motor action sequence in which worms switch from forward to backward crawling, turn, and then continue forward crawling. Their analysis showed that most variations ($\sim$65\%) in neural dynamics can be expressed by three principal components, and that neural dynamics in the resulting latent space trace cyclical trajectories on well-defined low dimensional manifolds corresponding to the motor action sequence (Figure S1). By identifying individual neurons, an experimental feat, these authors further determined that these topological structures in latent space were universally found among all five worms imaged in their study. 

\begin{figure}
\begin{center}
\includegraphics[scale=.16]{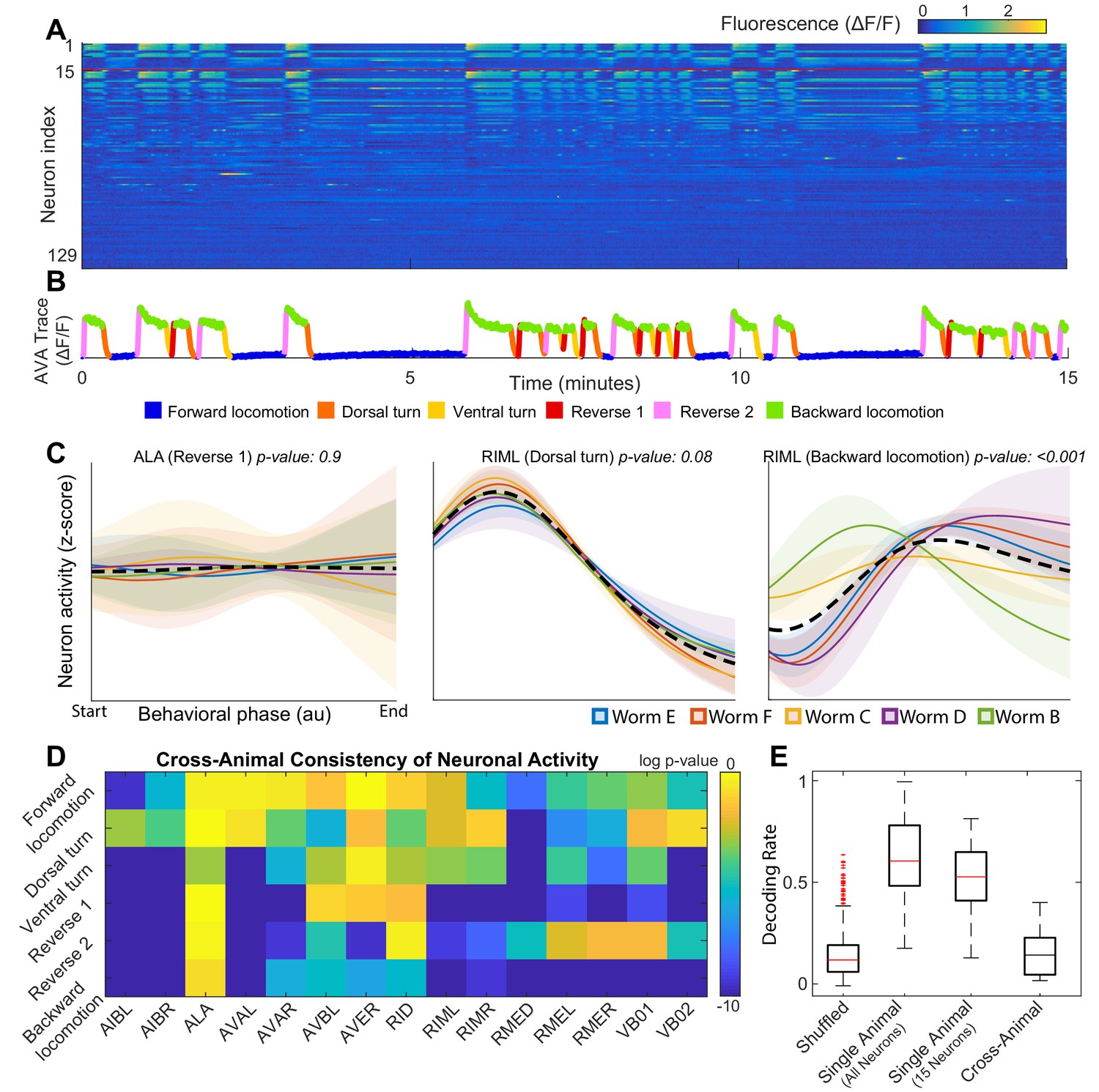}
\end{center}
\caption{\textbf{(A)} Calcium signals recorded in one animal for $\sim$15 minutes by \cite{kato}. Each row represents a single neuron. The top 15 rows (above the red line) correspond to neurons unambiguously identified in all animals (shared neurons). \textbf{(B)} Sample trace with corresponding behavioral state colored. \textbf{(C)} Neural dynamics of two neurons for specific behavior states. Colored solid lines are the mean activity for each animal, and the black dashed line is the mean activity for all animals. Shaded colored regions show 95\% confidence intervals. \textbf{(D)} Probabilities that neural dynamics from different individuals were drawn from the same distribution. \textbf{(E)} Attempt by \cite{brennan} to decode onset of backwards locomotion using neural dynamics for each animal and averaged neural dynamics across other four animals. Reproduced with permission from \cite{brennan}.}
\label{fig:figure1}
\end{figure}

Following this initial work \cite{kato}, the authors published several studies focusing on global organizational principles of C. elegans behavior \cite{nichols} \cite{hierarchical} \cite{energy}. Building on two of these works, \cite{brennan} found consistent differences between each individual’s neural dynamics, precluding the use of established dimensional reduction techniques. For example, among 15 neurons uniquely identified among all 5 worms, only 3 neurons displayed statistically consistent behavior  \hyperref[fig:figure1]{(Figure 1D)}. Examples of inconsistent behavior for unequivocally identified neurons (ALA and RIML) are shown in \hyperref[fig:figure1]{Figure 1C} where the average of ALA’s activity fails to resemble the behavior of any worm and where RIML’s activity is consistent among all animals during dorsal turns, but inconsistent during reverse crawling. Resulting from these discrepancies, topological structures identified by performing PCA on each worm's neural activity were no longer observed when data from all worms was pooled together.

\begin{figure}
\begin{center}
\includegraphics[scale=0.1]{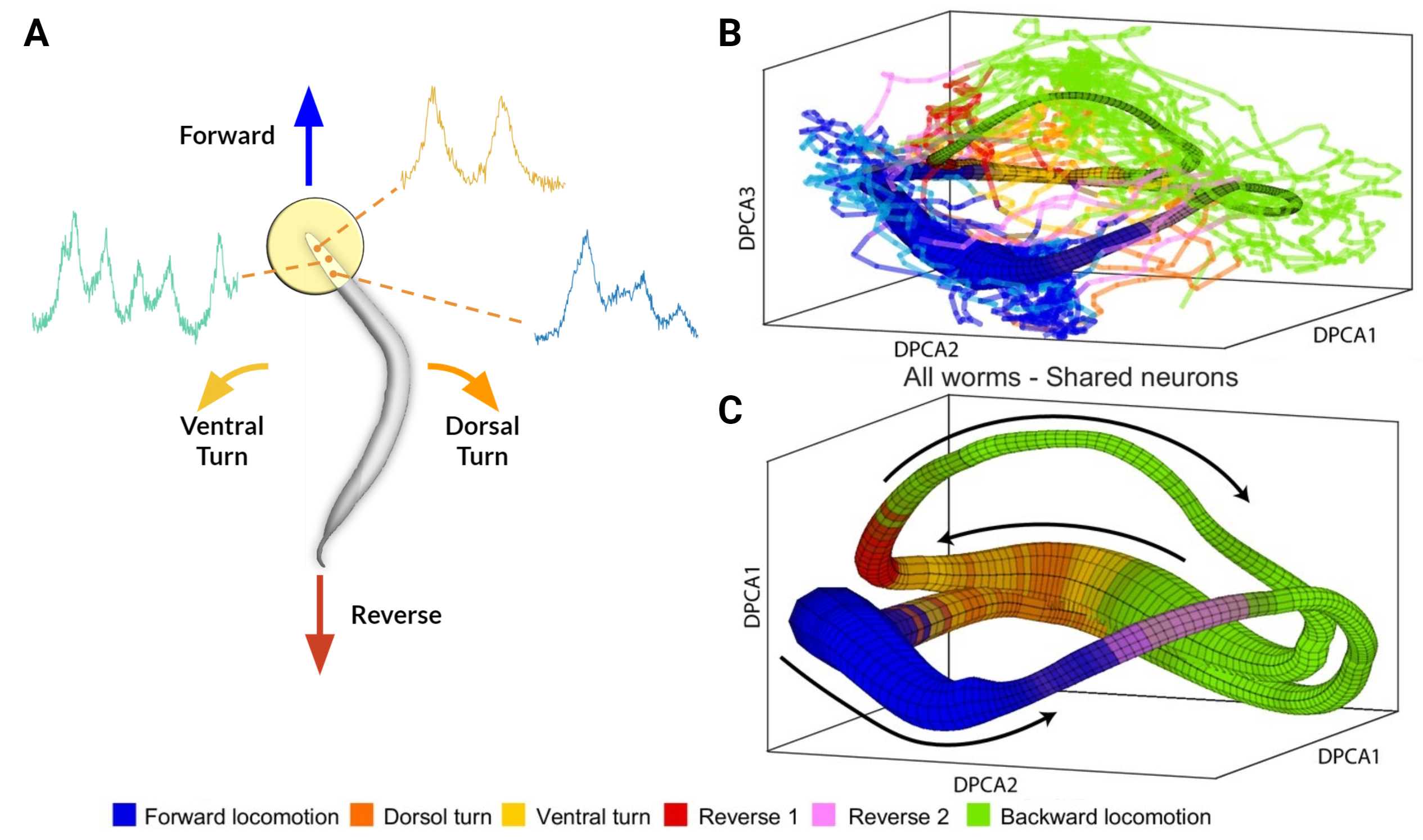}
\end{center}
\caption{\textbf{(A)} Rendering of calcium imaging experiment where activity of neurons in the head of the worm is recorded. Coloured arrows show main motor action behavioral states. \textbf{(B)} and \textbf{(C)} Resulting manifold from \cite{brennan}. \textbf{(B)} Manifold constructed from activity of four worms with coloured lines indicating neural activity of fifth worm. \textbf{(C)} Manifold constructed from neural activity of uniquely identified neurons ($n$=15) shared among all 5 worms. Black arrows correspond to cyclical transition of motor action sequence and colors correspond to motor action states. Modified with permission from \cite{brennan}.}
\label{fig:figure2}
\end{figure}

To address this issue, \cite{brennan} introduced a new algorithm, Asymmetric Diffusion Map Modeling (ADMM), which maps the neural activity of any worm to an universal manifold \hyperref[fig:figure2]{(Figure 2)}. To achieve this, ADMM first performs time-delay embedding of neural activity into phase space. Next, a transition probability matrix is constructed by calculating distances between points in phase space using a Gaussian kernel centered on the subsequent timestep. Finally, this asymmetric diffusion map is used to construct a manifold representative of neural activity. Contrasting conventional dimensional reduction techniques, ADMM allowed quantitative modeling by mapping neural activity from the manifold, and enabled the prediction of motor action states up to 30s ahead. Despite its success, the algorithm heavily relies on hyperparameters, such as embedding parameters, which are difficult to justify and tune. 

\subsection*{Graph Neural Networks}
Graph Neural Networks (GNNs) are a class of neural networks that explicitly use graph structure during computations through message passing algorithms where features are passed along edges between nodes and then aggregated for each node (\cite{og}; \cite{qchem}). These networks were inspired by the success of convolutional neural networks in the domain of two-dimensional image processing and failures when extending conventional convolutional networks to non-euclidean domains (\cite{gbrain}). In essence, because graphs can have arbitrary structure, the inductive bias of convolutional neural networks (equivariance to translational transformations \citep{equivariant}) often breaks down when applied to graphs. Addressing this issue, an early work on GNNs showed that one-hop message passing approximates spectral convolutions on graphs (\cite{gcn}). Subsequent works have examined the representational power of GNNs in relation to the Weisfeiler-Lehman isomorphism test \cite{GIN} and limitations of GNNs when learning graph moments (\cite{rep}).  From an applied perspective, GNNs have been widely successful in a wide variety of domains including relational inference (\cite{nri}; \cite{amortized}; \cite{entangled}), node classification \cite{gcn} \cite{graphsage}), point cloud segmentation \citep{pointcloud}, and traffic forecasting (\cite{spatiotraffic} \cite{diffusiontraffic}. In neuroscience, GNNs have been used on various tasks such as annotating cognitive state \citep{gnnfmri}, and several frameworks based on graph neural networks have been proposed for analyzing fMRI data (\cite{framework1}; \cite{framework2}).

\subsection*{Relational Inference}
Relational inference remains a longstanding challenge with early works in neuroscience seeking to quantify correlations between neurons \cite{granger}. Modern approaches to relational inference employ graph neural networks as their explicit reliance on graph structure forms a relational inductive bias \cite{intnetworks} \cite{gbrain}. In particular, our model is inspired by the Neural Relational Inference model (NRI) which uses a variational autoencoder for generating edges and a decoder for predicting trajectories of each object in a system \cite{nri}. By inferring edges, the NRI model explicitly captures interactions between objects and leverages the resulting graph as an inductive bias for various machine learning tasks. This model was successfully used to predict the trajectories of coupled Kuramoto oscillators, particles connected by springs, the pick and roll play from basketball, and motion capture visualizations. Subsequently, the authors developed Amortized Causal Discovery, a framework based on the NRI model which infers causal relations from time-dependent data \cite{amortized}.

\subsection*{Deep Learning in Neuroscience}
With the success of convolutional neural networks, researchers successfully applied deep learning to numerous domains in neuroscience (\cite{roleNN}) including MRI imaging \cite{dnnmedimaging} and connectomes \cite{mlconnectome} where algorithms can predict disorders such as autism \cite{autism}. Similarly, brain-computer interfaces (BCI) are a well-studied field related to our work as they focus on decoding macroscopic variables from measurements of neural activity \cite{Silva}. These studies generally involve fMRI or EEG data, which characterize neural activity on a population level, to varying amounts of success \cite{bci1} \cite{bci2} \cite{bci3} \cite{bci4}. Regardless, a challenge for the field is developing generalizable algorithms to individuals unseen during training \cite{survey-bci-dnn}.

\section*{Model}

In this section, we first present the general framework of our behavioral state classification and trajectory prediction models. Next, we detail the implementation of our neural network modules.

\subsection*{Framework}
We define the set of trajectories (calcium imaging traces) for each worm as $\displaystyle \mX_{\alpha} = \{\vx_1, ..., \vx_n\}$
where $\alpha$ denotes the label of the individual, $n$ the name of the neuron, and $\vx_n$ the feature vector of the neuron. In our case, $\vx_n$ corresponds to time-dependent normalized calcium traces and their derivatives for each neuron. Likewise, $x_n^t$ corresponds to the feature(s) of neuron $n$ at timestep $t$. Finally, the behavioral states of an individual are encoded as $\displaystyle \va_{\alpha} = (a^1,...,a^t)$ where a behavioral state $a$ is assigned for each timestep $t$.

\begin{figure}
\begin{center}
\includegraphics[scale=0.13]{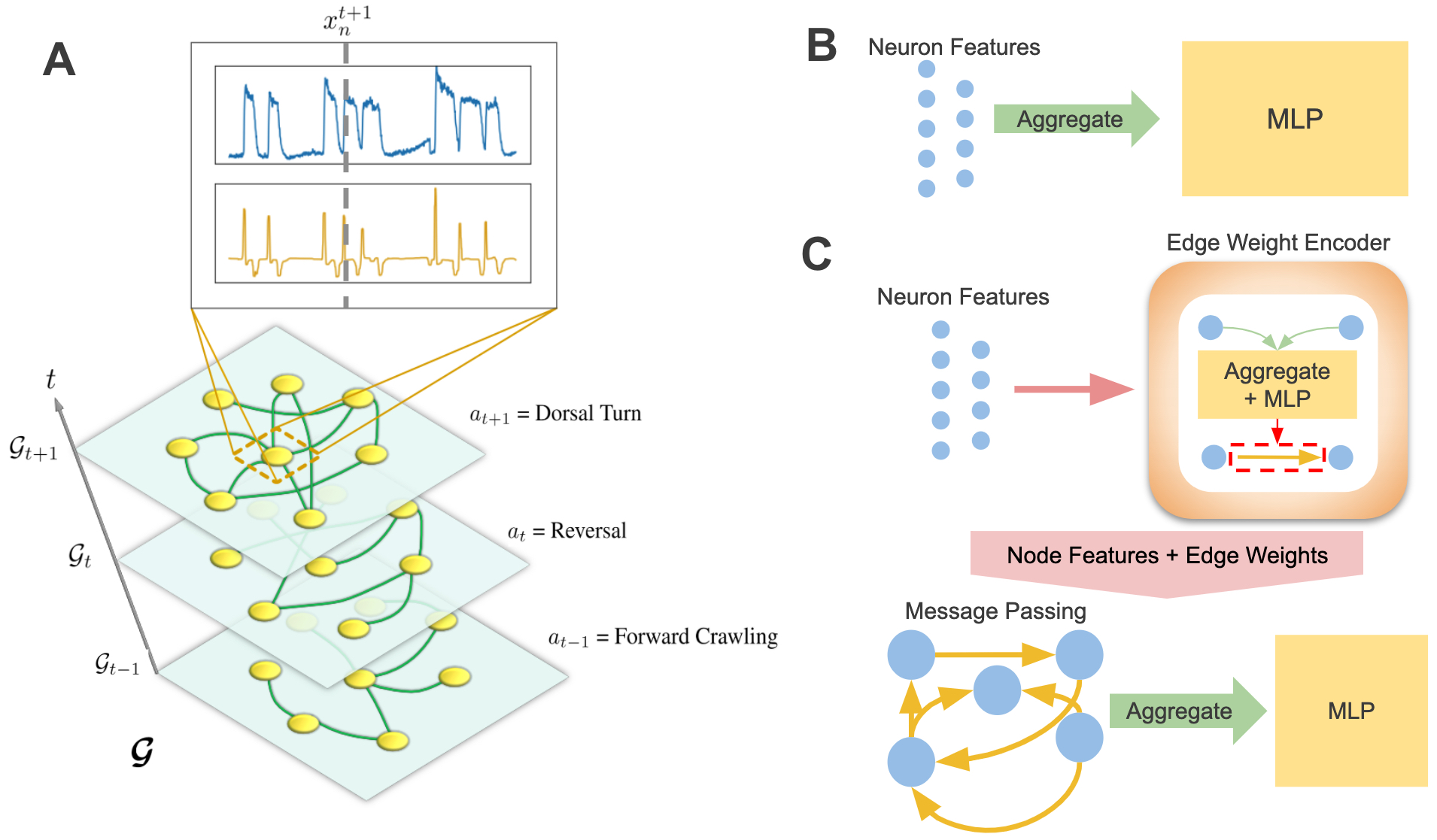}
\end{center}
\caption{\textbf{(A)} Visualization of temporal graph. Inset shows $\vx_n$ plotted against $t$ where the top is the calcium trace, and the bottom is its derivative. The dashed line intercepts the feature vectors at $t'$ = $t+1$ and denotes $x_n^{t+1}$. \textbf{(B)} and \textbf{(C)} are simplified visualizations of the MLP and GNN modules respectively.}
\label{fig:figure3}
\end{figure}

Separate models were developed for each task: behavioral state classification and trajectory prediction. In both cases, data from a worm $\alpha$ is structured as a temporal graph $\bm{\mathcal G_{\alpha}} = (\gG_{\alpha}^1,...,\gG_{\alpha}^t)$ \hyperref[fig:figure3]{(Figure 3A)} where each timestep is represented by a static graph whose nodes correspond to neurons. Following the notation above, the trajectories of each neuron's calcium traces are encoded as node features $\displaystyle \vx_n$, and the behavioral state of an individual is interpreted as a graph feature $\displaystyle a_{\alpha}^t$. For behavioral state classification, our model consists of the following (we omit $\alpha$ and $t$ in intermediary steps to simplify notation):
\begin{equation} \displaystyle \mH = f(\mX_{\alpha}^t)\end{equation}
\begin{equation} \displaystyle \vp = Softmax(\mH)\end{equation}
\begin{equation} \displaystyle \hat{a}_{\alpha}^t = Max(\vp) \end{equation}
where $\displaystyle f$ is an universal approximator/neural network module (described in the next section), $\mH$ are hidden features, $\vp$ is the probability that a system is in one of $k$ states, and $\hat{a}_{\alpha}^t$ is the most probable/predicted state.

For trajectory prediction, we developed a Markovian model for inferring trajectories of a consecutive timestep:
\begin{equation} \displaystyle \mH = f(\mX_{\alpha}^t)\end{equation}
\begin{equation} \displaystyle \mX_{\alpha}^{t+1} = \mX^t + \mH\end{equation}
where $\mH$ and $f$ are the same as before. We also experimented with non-Markovian models (RNNs) for which a hidden state is included for each timestep.

The structure of our models allows us to substitute various modules for $f$. While we include results from several neural networks, we focus on two representative models: a multi-layer perceptron (MLP) agnostic to graph structure \hyperref[fig:figure3]{(Figure 3B)} and a graph neural network (GNN) which explicitly computes on an inferred graph \hyperref[fig:figure3]{(Figure 3C)}.

\subsection*{Neural Network Modules: MLP and GNN}

Our MLP module aggregates the features of a graph and feeds the aggregated features into a two-layer MLP neural network:

\begin{equation} \displaystyle \mH = Aggregation(x_1^t,...,x_n^t)\end{equation}
\begin{equation} \displaystyle \mH_{out}^t = g(\mH)\end{equation}
where $g$ is a MLP. Contrasting the MLP module, our GNN relies on message passing between connected nodes and contains an encoder for edge weights $w_{ij}$:

\begin{equation} \displaystyle \mH_1 = g_{enc}(\mX^t)\end{equation}
\begin{equation} \displaystyle \label{eq:edges} \mH_{ij} = g(Aggregation(h_i,h_j))\end{equation}
\begin{equation} \displaystyle \vp_{ij} = Softmax(h_i, h_j)\end{equation}
\begin{equation} \displaystyle w_{ij}^t = p_{ij}^2\end{equation}
where \hyperref[eq:edges]{(9)} encodes a hidden representation $\mH_{ij}$ for the edges. Applying the softmax function to $\mH_{ij}$ produces a two dimensional probability vector normalized to 1. We define the second dimension $p_{ij}^2$ as the weight $w_{ij}$ of an edge between nodes $i$ and $j$. The edge weights either dynamically change in each timestep's inferred graph $\gG^t$ or remain fixed for the whole temporal graph $\bm{\mathcal G}$ of an individual worm. If the edges are static for the temporal graph, the aggregation step in \hyperref[eq:edges]{(9)} also averages hidden features across all timesteps.

After edges are encoded, the GNN performs a message passing and aggregation step:

\begin{equation} \displaystyle \label{eq:12} \mH_i = \sum_{j}^N w_{ij}^t x_j^t\end{equation}
\begin{equation} \displaystyle \mH_{out}^t = g(Aggregation(\mH_i))\end{equation}
The sum is performed over all nodes in the graph such that weighted messages are passed between connected nodes and potentially along self edges. The message passing step \hyperref[eq:12]{(12)} can also be formulated in terms of an inferred weighted adjacency matrix $\mA^t$ and node features $\mX^t$:
\begin{equation} \label{eq:msgpass} \displaystyle \mH^t = \mA^t \mX^t \end{equation}
Theoretically, an arbitrary number of message passing steps can be implemented; however, we did not find any improvements when using more than one step. In addition, we find that performance improves when using concatenation instead of summation during the aggregation step.

\section*{Experiments}

\subsection*{Data}
Our experiments were performed with data acquired in \cite{kato} and \cite{nichols}. We summarize various details about the data in this section; however, we direct the reader to each respective publication for specific experimental details.

\subsubsection*{Calcium Imaging}
Kato \emph{et. al.} \cite{kato} showed that neural activity corresponding to the motor action sequence lives on low dimensional manifolds. To record neuron level dynamics, they did whole-brain genetically encoded $\text{Ca}^{2+}$ imaging with single-cell-resolution and measured $\sim$100 neurons for around 18 minutes. They then normalized each calcium trace by peak fluorescence and identified neurons using spatial position and previous literature \citep{wormatlas}. Aside from imaging freely moving worms, the authors also examined robustness of topological features to sensory stimuli changes, hub neuron silencing, and immobilization. For simplicity, we limited our experiments to data collected on freely moving worms.

Nichols et. al. \cite{nichols} focused on differences in neural activity of C. elegans while awake or asleep and studied two different strains of worms, n2 ($n$=11) and npr1 ($n$=10). Because experiments in both studies were performed by the same group, most experimental procedures were similar, allowing us to easily process data to match the Kato dataset. While this dataset includes imaging data of each worm during quiescence, for consistency with the Kato dataset, we only included data before sleep was induced. Furthermore, we combined results for both strains of worms as we did not notice any statistically relevant differences between them.

\subsubsection*{Data Processing}
We normalized the calcium trace and its derivative of each neuron to [0,1]. Normalization was performed for the entire recorded calcium trace of a worm instead of within each batch because the relative magnitudes of the traces have been found to contain graded information about the worm's behavioral state (eg. crawling speed). To create training batches, we separated each calcium trace of approximately 3000-4000 timesteps into batches of 8 timesteps where each timestep corresponds to roughly 1/3 of a second. We chose batch sizes of 8 timesteps because visualization of calcium traces showed that most local variations occur within this time frame. Moreover, 8 timesteps roughly corresponds to 3 seconds which is about the amount of time a worm needs to execute a behavioral change. Finally, the batches were shuffled before being divided into 10 folds later used for cross-validation, ensuring that each fold is representative across the whole dataset.

To compare with previous works, we performed our experiments on uniquely identified neurons between the datasets that we investigated. Identifying specific neurons is an experimental challenge, and as such, only a small fraction of neurons were unequivocally labeled. A total of 15 neurons were uniquely identified between all worms ($n$=5) measured in the Kato dataset: (AIBL, AIBR, ALA, AVAL, AVAR, AVBL, AVER, RID, RIML, RIMR, RMED, RMEL, RMER, VB01, VB02). In addition, the Nichols dataset contained data from 21 worms with 3 uniquely identified neurons shared among all worms in both datasets: (AIBR, AVAL, VB02).

\begin{table}
\caption{\label{tab:table1} Classification Accuracy of Forward and Reverse Crawling}
\begin{center}
\scalebox{0.65}{
    \begin{tabular}{ |c|c|c|c| }
     \hline
     & Training Set & Evaluation Set (Kato) & Evaluation Set (Nichols)\\ 
     \hline
     \cite{brennan} & 83 & 81 & --- \\ 
     SVM & 98.8 $\pm$ .4 & 82.8 $\pm$ 7.6 & 79.0 $\pm$ 11.7\\
     MLP & 99.3 $\pm$ .6 & 93.9 $\pm$ 10.3 & 88.9  $\pm$ 11.4 \\ 
     GNN (Connectome) & 99.5 $\pm$ .6 & 96.8 $\pm$ 4.3 & 85.5  $\pm$ 12.9 \\ 
     GNN & \textbf{99.5} $\boldsymbol{\pm}$ \textbf{.5} & \textbf{97.7 $\boldsymbol{\pm}$ 3.1} & \textbf{95.5  $\boldsymbol{\pm}$ 6.1} \\ 
     \hline
    \end{tabular}}
\end{center}
\end{table}

\subsection*{Results}
Following \cite{brennan}, we used data from \cite{kato} for training/evaluating our models and data from \cite{nichols} as an extended evaluation set. Because whole brain imaging is incredibly difficult, our datasets were relatively small. To address this, we experimented with data augmentation by combining data from multiple worms in the Kato dataset during model training. For all experiments, we performed 10-fold cross validation on all permutations of worms in our training set. More details, along with supplemental experiments, can be found in the Supplementary Information.

\subsubsection*{Behavioral State Classification}
Our first experiment compared the performance of our models to state-of-the-art results reported in \cite{brennan}. Specifically, this experiment involved the classification of only two motor action states, forward and reverse crawling. Along with our models described above, we also experimented with a support vector machine (SVM) and a GNN which computes with edges derived from the physical connectome \citep{white}. In particular, we incorporated the connectome into our model to investigate whether physical/structural connections between neurons can serve as a favourable inductive bias for our GNN. Our results are shown in \hyperref[tab:table1]{Table 1} where Training Set denotes test set accuracy after training on the same worm and Evaluation Set denotes evaluation/generalization accuracy on worms unseen during training. 

Our deep learning models clearly outperformed the SVM and state-of-the-art results, demonstrating the ability of our models to successfully classify behavioral states and generalize to other worms. Interestingly, the SVM matched the performance of our deep learning models on test set accuracy; however, its generalization performance on unseen individuals was significantly worse than our deep learning models. As such, the SVM distinctly illustrates challenges of individual variability for model development in neural systems despite the simplicity of our experiments which involve the same set of unequivocally identified neurons. Similarly, our GNN using edges derived from the connectome performed well on the test set but generalized worse than when using inferred edges. We hypothesize that the detrimental effect of using the connectome may be attributed to the model's lack of expressiveness and the distinction between inferred/functional and structural connectivity (See S.1.4.3).

\begin{figure}
\begin{center}
\includegraphics[scale=0.11]{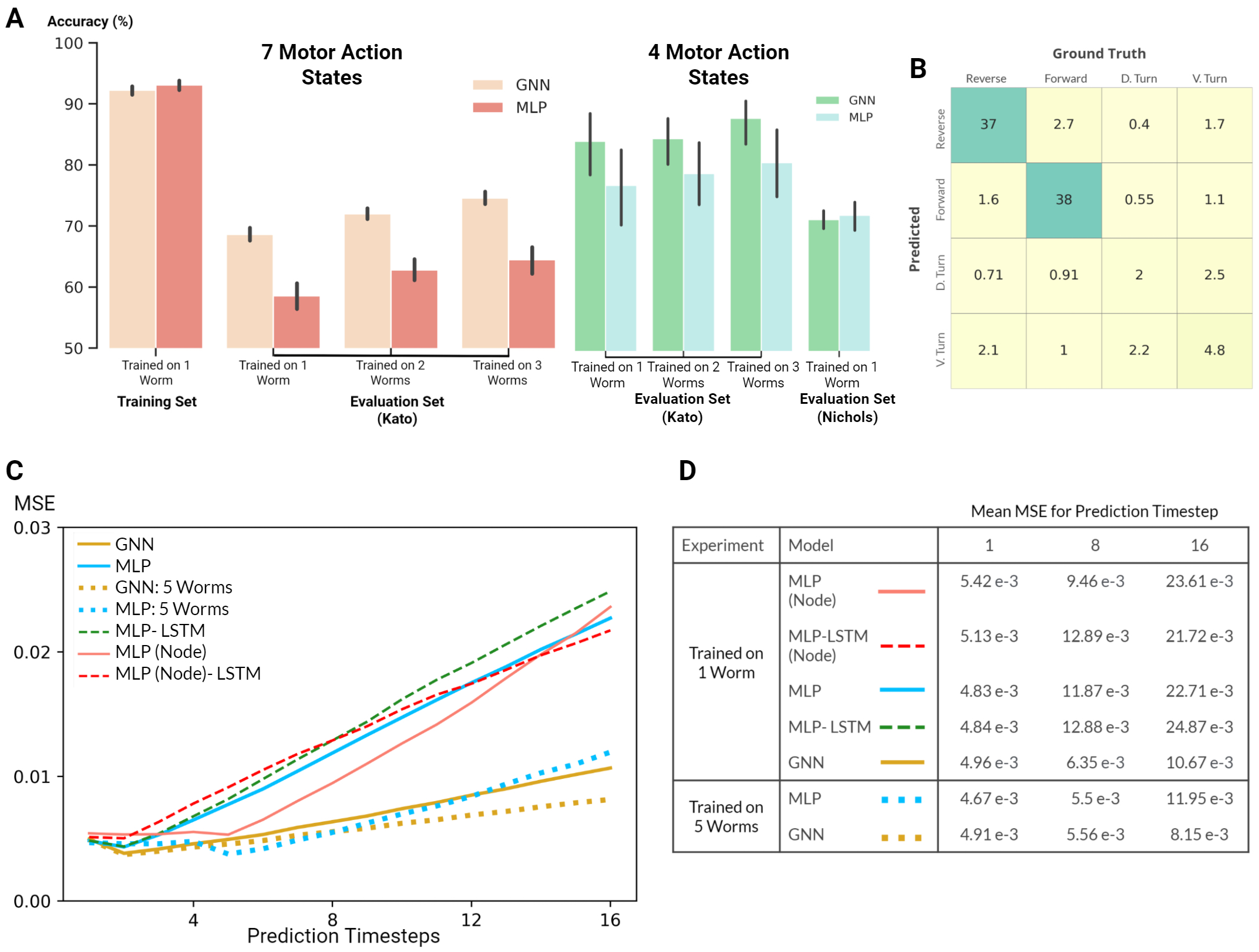}
\end{center}
\caption{\textbf{(A)} Classification accuracy of our GNN and MLP models where black vertical lines show statistical spread. \textbf{Left}: Classification of 7 motor action states within the Kato dataset. \textbf{Right}: Classification of 4 motor action states on both the Kato and Nichols datasets. \textbf{(B)} Confusion matrix. Percent occurrence of predicted states against labeled states when evaluating on the Nichols dataset. \textbf{(C)} Mean squared error (MSE) of the GNN and various MLP models evaluated on the Nichols dataset. All models were trained using data from one worm or five worms in the Kato Dataset. \textbf{(D)} Table of MSE values for all models for 1, 8, and 16 timesteps.}
\label{fig:figure4}
\end{figure}

Following the previous experiment, we applied our MLP and GNN models to the harder task of classifying all behavioral states labeled in the Kato dataset \hyperref[fig:figure4]{(Figure 4A)}. Within this dataset, 7 states were labeled: Forward Crawling, Forward Slowing, Reverse 1, Reverse 2, Sustained Reverse Crawling, Dorsal Turn, and Ventral Turn. In comparison to the Kato dataset, only 4 states were labeled in the Nichols dataset: reverse crawling, forward crawling, ventral turn, and dorsal turn. For compatibility, we mapped the 7 states of the Kato dataset to 4 states of the Nichols dataset when using the Nichols dataset as an extended evaluation set.

Despite the harder task of classifying 7 states, our models achieved a classification accuracy of $\sim$92\% on the same worm \hyperref[fig:figure4]{(Figure 4A: Left)}. Moreover, our GNN trained on three worms in the Kato dataset generalized with an accuracy of 87\% \hyperref[fig:figure4]{(Figure 4A: Right)} when classifying 4 states on the remaining unseen worms. This substantially exceeds the performance of our MLP model and \cite{brennan} who report a 81\% cross-animal accuracy on two states. Nevertheless, both MLP and GNN models generalized equally well ($\sim$70\%) to the 21 unseen worms of the Nichols dataset. These experiments consistently demonstrate that our GNN exceeds the performance of state-of-the-art techniques and also often exceeds the performance of our baseline MLP model. 

\subsubsection*{Neuron-level Trajectory Prediction}
For trajectory prediction, we predicted each neuron's calcium trace and its derivative (normalized to [0,1]) for 8 timesteps during training and 16 timesteps during evaluation/validation. While training our Markovian models, scheduled sampling was performed to minimize the accumulation of error \citep{scheduled}. In addition to our Markovian models, we also experimented with RNN implementations trained with burn-in periods of four timesteps. For evaluation, we averaged the loss per prediction timestep across all batches. Our experiments primarily focused on generalization performance of our models on the extended evaluation/Nichols dataset \hyperref[fig:figure4]{(Figure 4C)}.

Predicting neuron-level trajectory using deep learning is fairly novel since advances in whole-brain imaging are recent and limited to few organisms. Because calcium traces are notoriously noisy and our dataset is relatively small, the performance of our model is poor; however, inspecting the MSE as a function of prediction step \hyperref[fig:figure4]{(Figure 4C)} demonstrates that all deep learning models are able to learn transitions in the system. Moreover, increasing the number of worms included during training also improved generalization performance of our MLP and GNN models. Perhaps most surprising, our Markovian GNN outperformed all MLP models and their derived RNN variants. We attribute this result to the largely deterministic nature of neural dynamics, characterized by sparse bifurcations on the latent manifold, and the inductive bias of GNNs. As a result, given a single timestep, our GNN model was able to predict future trajectories on unseen worms for at least 16 timesteps and clearly outperformed all other models.

\section*{Discussion}
For both tasks, our GNN consistently matched or exceeded our MLP model which we accredit to its favourable inductive bias. Kato \emph{et. al.} \cite{kato} established that projecting neural dynamics onto three principal components for each worm reveals universal topological structures; however, attempts to project neural dynamics onto shared principal components of all worms failed to display any meaningful structure. Thus, variability in each worm's neural activity, corresponding to low dimensional manifolds in latent space, is represented by different linear combinations of neurons. In other words, relevant topological structures in latent space are loosely related by linear transformations of node features. We speculate that our GNN's performance stems from its explicit structure of message passing along inferred edges which is analogous to learning linear transformations of node features (see equation \hyperref[eq:msgpass]{(14)}).

Interestingly, our model’s performance was not significantly impacted by using 3 neurons ($\sim$1\% of all neurons) instead of 15 ($\sim$5\% of all neurons). This is not surprising because neurons strongly coupled to the motor action sequence retain most information \citep{coupledneuron}, a fact consistent with \cite{brennan} who found that strategically choosing 1 neuron retains $\sim$75\% of the information contained in the larger set of 15 neurons.
 
Finally, as a critical question, we ask whether our model’s performance stems from choosing a stereotyped organism that is well studied and biologically simple, or if our results imply a path towards generalizable/universal machine learning in neural systems. While the neurophysiology of C. elegans is quite complex, the motor action sequence we studied is relatively simple, especially in comparison to other organisms and cognitive functions. Moreover, organisms are adaptive and capable of learning new behavior, a fact not represented in our dataset. However, a recent astounding study \cite{monkeys} measured neural dynamics in monkeys trained to perform action sequences and determined that learned latent dynamics live in low-dimensional manifolds that were conserved throughout the length of the study. By aligning latent dynamics, their model accurately decoded the action of monkeys up to two years after the model was trained despite changes in biology (eg. neuron turnover, adaptation to implants). Consequently, we posit that techniques similar to those used in our model may broadly apply to more complex organisms and functions.

\section*{Conclusion}
In this study, we examined the ability of neural networks to classify higher-order function and predict neuron level dynamics. In addition, inspired by global organizational principles of behavior discovered in previous studies, we demonstrated the ability of neural networks to generalize to unseen organisms. Specifically, our models exceeded the performance of previous studies in behavioral state classification of C. elegans. Furthermore, our models successfully generalized to unseen organisms, both within the same study, and in a separate experiment spaced years apart. We found that a simple MLP performs remarkably well on unseen organisms. Nevertheless, our graph neural network, which explicitly learns linear transformations of node features, matched or exceeded the performance of graph agnostic models in all experiments. 

We note that our results of generalization on both higher-order functions and neuron-level dynamics (macroscopic and microscopic) suggests wide applicability of our technique to numerous machine learning tasks in neuroscience and hierarchical dynamical systems. A promising research direction is the hierarchical relationship between neuron-level and population-level dynamics. Breakthroughs in this direction may inform machine learning models working with population-level functional and imaging techniques, such as EEG or fMRI, which are readily available and widespread. In addition, in this study, we only focused on simple machine learning tasks and imaging data taken under similar experimental conditions. Further studies may involve more complex tasks such as those involving graded information in neural dynamics, changes in sensory stimuli, acquisition of learned behaviors, and higher-order functions comprised of complicated sequences of behavior. From a machine learning perspective, the development of a recurrent graph neural network with a suitable attention kernel may greatly aid model performance. Moreover, additional work is needed in examining and improving model performance on arbitrary sets of neurons as neuron identification is experimentally challenging and limited to small systems.  Finally, our results show that data augmentation through the inclusion of more individuals can significantly improve generalization performance in microscopic neural systems.

\section*{Conflict of Interest Statement}
The authors declare that the research was conducted in the absence of any commercial or financial relationships that could be construed as a potential conflict of interest.

\section*{Author Contributions}
Experiments and models were conceived by P.Y. Wang. S. Sapra assisted with the implementation of various algorithms. The manuscript was written and revised after numerous iterations by all the authors.

\section*{Funding}
This work was supported by unrestricted funds to the Center for Engineered Natural Intelligence. 

\section*{Acknowledgments}
The authors thank the authors of \cite{brennan} for graciously allowing reproductions of their figures. In addition, the authors thank the Zimmer Lab for making their data available online (data from \cite{kato} and \cite{nichols} can be found \href{https://osf.io/a64uz/}{here}). P.Y. Wang is grateful to Ilya Valmianski for insightful discussion and guidance.

\section*{Data Availability Statement}
The datasets \cite{kato} and \cite{nichols} analyzed for this study can be found in the OSF repository \href{https://osf.io/a64uz/}{here}.

\section*{Bibliography}
\bibliography{Wang}

\begin{thebibliography}{56}
\providecommand{\natexlab}[1]{#1}
\providecommand{\url}[1]{\texttt{#1}}
\expandafter\ifx\csname urlstyle\endcsname\relax
  \providecommand{\doi}[1]{doi: #1}\else
  \providecommand{\doi}{doi: \begingroup \urlstyle{rm}\Url}\fi

\bibitem[Golowasch et~al.(2002)Golowasch, Goldman, Abbott, and Marder]{failure}
Jorge Golowasch, Mark~S Goldman, LF~Abbott, and Eve Marder.
\newblock Failure of averaging in the construction of a conductance-based
  neuron model.
\newblock \emph{Journal of neurophysiology}, 87\penalty0 (2):\penalty0
  1129--1131, 2002.

\bibitem[Prinz et~al.(2004)Prinz, Bucher, and Marder]{similar}
Astrid~A Prinz, Dirk Bucher, and Eve Marder.
\newblock Similar network activity from disparate circuit parameters.
\newblock \emph{Nature neuroscience}, 7\penalty0 (12):\penalty0 1345--1352,
  2004.

\bibitem[Fr{\'e}gnac(2017)]{fregnac2017big}
Yves Fr{\'e}gnac.
\newblock Big data and the industrialization of neuroscience: A safe roadmap
  for understanding the brain?
\newblock \emph{Science}, 358\penalty0 (6362):\penalty0 470--477, 2017.

\bibitem[Churchland et~al.(2010)Churchland, Cunningham, Kaufman, Ryu, and
  Shenoy]{churchland2010cortical}
Mark~M Churchland, John~P Cunningham, Matthew~T Kaufman, Stephen~I Ryu, and
  Krishna~V Shenoy.
\newblock Cortical preparatory activity: representation of movement or first
  cog in a dynamical machine?
\newblock \emph{Neuron}, 68\penalty0 (3):\penalty0 387--400, 2010.

\bibitem[Goldman et~al.(2001)Goldman, Golowasch, Marder, and
  Abbott]{robustness-global-structure}
{Mark S} Goldman, Jorge Golowasch, Eve Marder, and {L. F.} Abbott.
\newblock Global structure, robustness, and modulation of neuronal models.
\newblock \emph{Journal of Neuroscience}, 21\penalty0 (14):\penalty0
  5229--5238, 2001.

\bibitem[Prevedel et~al.(2014)Prevedel, Yoon, Hoffmann, Pak, Wetzstein, Kato,
  Schr{\"o}del, Raskar, Zimmer, Boyden, et~al.]{sim-imaging}
Robert Prevedel, Young-Gyu Yoon, Maximilian Hoffmann, Nikita Pak, Gordon
  Wetzstein, Saul Kato, Tina Schr{\"o}del, Ramesh Raskar, Manuel Zimmer,
  Edward~S Boyden, et~al.
\newblock Simultaneous whole-animal 3d imaging of neuronal activity using
  light-field microscopy.
\newblock \emph{Nature methods}, 11\penalty0 (7):\penalty0 727--730, 2014.

\bibitem[Kato et~al.(2015)Kato, Kaplan, Schr{\"o}del, Skora, Lindsay, Yemini,
  Lockery, and Zimmer]{kato}
Saul Kato, Harris~S Kaplan, Tina Schr{\"o}del, Susanne Skora, Theodore~H
  Lindsay, Eviatar Yemini, Shawn Lockery, and Manuel Zimmer.
\newblock Global brain dynamics embed the motor command sequence of
  caenorhabditis elegans.
\newblock \emph{Cell}, 163\penalty0 (3):\penalty0 656--669, 2015.

\bibitem[White et~al.(1986)White, Southgate, Thomson, and Brenner]{white}
John~G White, Eileen Southgate, J~Nichol Thomson, and Sydney Brenner.
\newblock The structure of the nervous system of the nematode caenorhabditis
  elegans.
\newblock \emph{Philos Trans R Soc Lond B Biol Sci}, 314\penalty0
  (1165):\penalty0 1--340, 1986.

\bibitem[Bargmann and Marder(2013)]{bargmann2013connectome}
Cornelia~I Bargmann and Eve Marder.
\newblock From the connectome to brain function.
\newblock \emph{Nature methods}, 10\penalty0 (6):\penalty0 483, 2013.

\bibitem[Varshney et~al.(2011)Varshney, Chen, Paniagua, Hall, and
  Chklovskii]{varshney2011structural}
Lav~R Varshney, Beth~L Chen, Eric Paniagua, David~H Hall, and Dmitri~B
  Chklovskii.
\newblock Structural properties of the caenorhabditis elegans neuronal network.
\newblock \emph{PLoS Comput Biol}, 7\penalty0 (2):\penalty0 e1001066, 2011.

\bibitem[Cook et~al.(2019)Cook, Jarrell, Brittin, Wang, Bloniarz, Yakovlev,
  Nguyen, Tang, Bayer, Duerr, et~al.]{connectome}
Steven~J Cook, Travis~A Jarrell, Christopher~A Brittin, Yi~Wang, Adam~E
  Bloniarz, Maksim~A Yakovlev, Ken~CQ Nguyen, Leo T-H Tang, Emily~A Bayer,
  Janet~S Duerr, et~al.
\newblock Whole-animal connectomes of both caenorhabditis elegans sexes.
\newblock \emph{Nature}, 571\penalty0 (7763):\penalty0 63--71, 2019.

\bibitem[Wen and Kimura(2020)]{wba}
Chentao Wen and Koutarou~D Kimura.
\newblock How do we know how the brain works?—analyzing whole brain
  activities with classic mathematical and machine learning methods.
\newblock \emph{Japanese Journal of Applied Physics}, 59\penalty0 (3):\penalty0
  030501, 2020.

\bibitem[Sarma et~al.(2018)Sarma, Lee, Portegys, Ghayoomie, Jacobs, Alicea,
  Cantarelli, Currie, Gerkin, Gingell, et~al.]{openworm}
Gopal~P Sarma, Chee~Wai Lee, Tom Portegys, Vahid Ghayoomie, Travis Jacobs,
  Bradly Alicea, Matteo Cantarelli, Michael Currie, Richard~C Gerkin, Shane
  Gingell, et~al.
\newblock Openworm: overview and recent advances in integrative biological
  simulation of caenorhabditis elegans.
\newblock \emph{Philosophical Transactions of the Royal Society B},
  373\penalty0 (1758):\penalty0 20170382, 2018.

\bibitem[Gleeson et~al.(2018)Gleeson, Lung, Grosu, Hasani, and Larson]{c302}
Padraig Gleeson, David Lung, Radu Grosu, Ramin Hasani, and Stephen~D Larson.
\newblock c302: a multiscale framework for modelling the nervous system of
  caenorhabditis elegans.
\newblock \emph{Philosophical Transactions of the Royal Society B: Biological
  Sciences}, 373\penalty0 (1758):\penalty0 20170379, 2018.

\bibitem[Nichols et~al.(2017)Nichols, Eichler, Latham, and Zimmer]{nichols}
Annika~LA Nichols, Tom{\'a}{\v{s}} Eichler, Richard Latham, and Manuel Zimmer.
\newblock A global brain state underlies c. elegans sleep behavior.
\newblock \emph{Science}, 356\penalty0 (6344), 2017.

\bibitem[Kaplan et~al.(2020)Kaplan, Thula, Khoss, and Zimmer]{hierarchical}
Harris~S Kaplan, Oriana~Salazar Thula, Niklas Khoss, and Manuel Zimmer.
\newblock Nested neuronal dynamics orchestrate a behavioral hierarchy across
  timescales.
\newblock \emph{Neuron}, 105\penalty0 (3):\penalty0 562--576, 2020.

\bibitem[Skora et~al.(2018)Skora, Mende, and Zimmer]{energy}
Susanne Skora, Fanny Mende, and Manuel Zimmer.
\newblock Energy scarcity promotes a brain-wide sleep state modulated by
  insulin signaling in c. elegans.
\newblock \emph{Cell reports}, 22\penalty0 (4):\penalty0 953--966, 2018.

\bibitem[Brennan and Proekt(2019)]{brennan}
Connor Brennan and Alexander Proekt.
\newblock A quantitative model of conserved macroscopic dynamics predicts
  future motor commands.
\newblock \emph{Elife}, 8:\penalty0 e46814, 2019.

\bibitem[{Scarselli} et~al.(2009){Scarselli}, {Gori}, {Tsoi}, {Hagenbuchner},
  and {Monfardini}]{og}
F.~{Scarselli}, M.~{Gori}, A.~C. {Tsoi}, M.~{Hagenbuchner}, and
  G.~{Monfardini}.
\newblock The graph neural network model.
\newblock \emph{IEEE Transactions on Neural Networks}, 20\penalty0
  (1):\penalty0 61--80, 2009.

\bibitem[Gilmer et~al.(2017)Gilmer, Schoenholz, Riley, Vinyals, and
  Dahl]{qchem}
Justin Gilmer, Samuel~S Schoenholz, Patrick~F Riley, Oriol Vinyals, and
  George~E Dahl.
\newblock Neural message passing for quantum chemistry.
\newblock In \emph{Proceedings of the 34th International Conference on Machine
  Learning-Volume 70}, pages 1263--1272, 2017.

\bibitem[Battaglia et~al.(2018)Battaglia, Hamrick, Bapst, Sanchez, Zambaldi,
  Malinowski, Tacchetti, Raposo, Santoro, Faulkner, Gulcehre, Song, Ballard,
  Gilmer, Dahl, Vaswani, Allen, Nash, Langston, Dyer, Heess, Wierstra, Kohli,
  Botvinick, Vinyals, Li, and Pascanu]{gbrain}
Peter Battaglia, Jessica Blake~Chandler Hamrick, Victor Bapst, Alvaro Sanchez,
  Vinicius Zambaldi, Mateusz Malinowski, Andrea Tacchetti, David Raposo, Adam
  Santoro, Ryan Faulkner, Caglar Gulcehre, Francis Song, Andy Ballard, Justin
  Gilmer, George~E. Dahl, Ashish Vaswani, Kelsey Allen, Charles Nash,
  Victoria~Jayne Langston, Chris Dyer, Nicolas Heess, Daan Wierstra, Pushmeet
  Kohli, Matt Botvinick, Oriol Vinyals, Yujia Li, and Razvan Pascanu.
\newblock Relational inductive biases, deep learning, and graph networks.
\newblock \emph{arXiv}, 2018.

\bibitem[Cohen and Welling(2016)]{equivariant}
Taco Cohen and Max Welling.
\newblock Group equivariant convolutional networks.
\newblock In \emph{International conference on machine learning}, pages
  2990--2999, 2016.

\bibitem[Kipf and Welling(2016)]{gcn}
Thomas~N Kipf and Max Welling.
\newblock Semi-supervised classification with graph convolutional networks.
\newblock \emph{arXiv preprint arXiv:1609.02907}, 2016.

\bibitem[Xu et~al.(2018)Xu, Hu, Leskovec, and Jegelka]{GIN}
Keyulu Xu, Weihua Hu, Jure Leskovec, and Stefanie Jegelka.
\newblock How powerful are graph neural networks?
\newblock \emph{arXiv preprint arXiv:1810.00826}, 2018.

\bibitem[Dehmamy et~al.(2019)Dehmamy, Barab{\'a}si, and Yu]{rep}
Nima Dehmamy, Albert-L{\'a}szl{\'o} Barab{\'a}si, and Rose Yu.
\newblock Understanding the representation power of graph neural networks in
  learning graph topology.
\newblock In \emph{Advances in Neural Information Processing Systems}, pages
  15413--15423, 2019.

\bibitem[Kipf et~al.(2018)Kipf, Fetaya, Wang, Welling, and Zemel]{nri}
Thomas Kipf, Ethan Fetaya, Kuan-Chieh Wang, Max Welling, and Richard Zemel.
\newblock Neural relational inference for interacting systems.
\newblock In \emph{International Conference on Machine Learning}, pages
  2688--2697, 2018.

\bibitem[Löwe et~al.(2020)Löwe, Madras, Zemel, and Welling]{amortized}
Sindy Löwe, David Madras, Richard Zemel, and Max Welling.
\newblock Amortized causal discovery: Learning to infer causal graphs from
  time-series data, 2020.

\bibitem[Raposo et~al.(2017)Raposo, Santoro, Barrett, Pascanu, Lillicrap, and
  Battaglia]{entangled}
David Raposo, Adam Santoro, David Barrett, Razvan Pascanu, Timothy Lillicrap,
  and Peter Battaglia.
\newblock Discovering objects and their relations from entangled scene
  representations.
\newblock \emph{arXiv preprint arXiv:1702.05068}, 2017.

\bibitem[Hamilton et~al.(2017)Hamilton, Ying, and Leskovec]{graphsage}
Will Hamilton, Zhitao Ying, and Jure Leskovec.
\newblock Inductive representation learning on large graphs.
\newblock In \emph{Advances in neural information processing systems}, pages
  1024--1034, 2017.

\bibitem[Wang et~al.(2019)Wang, Sun, Liu, Sarma, Bronstein, and
  Solomon]{pointcloud}
Yue Wang, Yongbin Sun, Ziwei Liu, Sanjay~E Sarma, Michael~M Bronstein, and
  Justin~M Solomon.
\newblock Dynamic graph cnn for learning on point clouds.
\newblock \emph{Acm Transactions On Graphics (tog)}, 38\penalty0 (5):\penalty0
  1--12, 2019.

\bibitem[Yu et~al.(2017)Yu, Yin, and Zhu]{spatiotraffic}
Bing Yu, Haoteng Yin, and Zhanxing Zhu.
\newblock Spatio-temporal graph convolutional networks: A deep learning
  framework for traffic forecasting.
\newblock \emph{arXiv preprint arXiv:1709.04875}, 2017.

\bibitem[Li et~al.(2017)Li, Yu, Shahabi, and Liu]{diffusiontraffic}
Yaguang Li, Rose Yu, Cyrus Shahabi, and Yan Liu.
\newblock Diffusion convolutional recurrent neural network: Data-driven traffic
  forecasting.
\newblock \emph{arXiv preprint arXiv:1707.01926}, 2017.

\bibitem[Zhang and Bellec(2019)]{gnnfmri}
Yu~Zhang and Pierre Bellec.
\newblock Functional annotation of human cognitive states using graph
  convolution networks.
\newblock 2019.

\bibitem[Li and Duncan(2020)]{framework1}
Xiaoxiao Li and James Duncan.
\newblock Braingnn: Interpretable brain graph neural network for fmri analysis.
\newblock \emph{bioRxiv}, 2020.

\bibitem[Kim and Ye(2020)]{framework2}
Byung-Hoon Kim and Jong~Chul Ye.
\newblock Understanding graph isomorphism network for brain mr functional
  connectivity analysis.
\newblock \emph{arXiv preprint arXiv:2001.03690}, 2020.

\bibitem[Granger(1969)]{granger}
Clive~WJ Granger.
\newblock Investigating causal relations by econometric models and
  cross-spectral methods.
\newblock \emph{Econometrica: journal of the Econometric Society}, pages
  424--438, 1969.

\bibitem[Battaglia et~al.(2016)Battaglia, Pascanu, Lai, Rezende,
  et~al.]{intnetworks}
Peter Battaglia, Razvan Pascanu, Matthew Lai, Danilo~Jimenez Rezende, et~al.
\newblock Interaction networks for learning about objects, relations and
  physics.
\newblock In \emph{Advances in neural information processing systems}, pages
  4502--4510, 2016.

\bibitem[Glaser et~al.(2019)Glaser, Benjamin, Farhoodi, and Kording]{roleNN}
Joshua~I Glaser, Ari~S Benjamin, Roozbeh Farhoodi, and Konrad~P Kording.
\newblock The roles of supervised machine learning in systems neuroscience.
\newblock \emph{Progress in neurobiology}, 175:\penalty0 126--137, 2019.

\bibitem[Lundervold and Lundervold(2019)]{dnnmedimaging}
Alexander~Selvikv{\aa}g Lundervold and Arvid Lundervold.
\newblock An overview of deep learning in medical imaging focusing on mri.
\newblock \emph{Zeitschrift f{\"u}r Medizinische Physik}, 29\penalty0
  (2):\penalty0 102--127, 2019.

\bibitem[Brown and Hamarneh(2016)]{mlconnectome}
Colin~J Brown and Ghassan Hamarneh.
\newblock Machine learning on human connectome data from mri.
\newblock \emph{arXiv preprint arXiv:1611.08699}, 2016.

\bibitem[Brown et~al.(2018)Brown, Kawahara, and Hamarneh]{autism}
Colin~J Brown, Jeremy Kawahara, and Ghassan Hamarneh.
\newblock Connectome priors in deep neural networks to predict autism.
\newblock In \emph{2018 IEEE 15th International Symposium on Biomedical Imaging
  (ISBI 2018)}, pages 110--113. IEEE, 2018.

\bibitem[Silva(2018)]{Silva}
Gabriel~A Silva.
\newblock {A New Frontier: The Convergence of Nanotechnology, Brain Machine
  Interfaces, and Artificial Intelligence}.
\newblock \emph{Frontiers in Neuroscience}, 12:\penalty0 843, 2018.
\newblock ISSN 1662-4548.
\newblock \doi{10.3389/fnins.2018.00843}.

\bibitem[Bashivan et~al.(2015)Bashivan, Rish, Yeasin, and Codella]{bci1}
Pouya Bashivan, Irina Rish, Mohammed Yeasin, and Noel Codella.
\newblock Learning representations from eeg with deep recurrent-convolutional
  neural networks.
\newblock \emph{arXiv preprint arXiv:1511.06448}, 2015.

\bibitem[Kwak et~al.(2017)Kwak, M{\"u}ller, and Lee]{bci2}
No-Sang Kwak, Klaus-Robert M{\"u}ller, and Seong-Whan Lee.
\newblock A convolutional neural network for steady state visual evoked
  potential classification under ambulatory environment.
\newblock \emph{PloS one}, 12\penalty0 (2):\penalty0 e0172578, 2017.

\bibitem[Mensch et~al.(2017)Mensch, Mairal, Bzdok, Thirion, and
  Varoquaux]{bci3}
Arthur Mensch, Julien Mairal, Danilo Bzdok, Bertrand Thirion, and Ga{\"e}l
  Varoquaux.
\newblock Learning neural representations of human cognition across many fmri
  studies.
\newblock In \emph{Advances in neural information processing systems}, pages
  5883--5893, 2017.

\bibitem[Makin et~al.(2020)Makin, Moses, and Chang]{bci4}
Joseph~G Makin, David~A Moses, and Edward~F Chang.
\newblock Machine translation of cortical activity to text with an
  encoder--decoder framework.
\newblock Technical report, Nature Publishing Group, 2020.

\bibitem[Zhang et~al.(2019)Zhang, Yao, Wang, Monaghan, Mcalpine, and
  Zhang]{survey-bci-dnn}
Xiang Zhang, Lina Yao, Xianzhi Wang, Jessica Monaghan, David Mcalpine, and
  Yu~Zhang.
\newblock A survey on deep learning based brain computer interface: Recent
  advances and new frontiers.
\newblock \emph{arXiv preprint arXiv:1905.04149}, 2019.

\bibitem[Altun et~al.(2002-2020)Altun, Herndon, Wolkow, Crocker, Lints, and
  Hall]{wormatlas}
Z.~F. Altun, L.~A. Herndon, C.~A. Wolkow, C.~Crocker, R.~Lints, and D.~H. Hall.
\newblock Worm atlas, 2002-2020.

\bibitem[Bengio et~al.(2015)Bengio, Vinyals, Jaitly, and Shazeer]{scheduled}
Samy Bengio, Oriol Vinyals, Navdeep Jaitly, and Noam Shazeer.
\newblock Scheduled sampling for sequence prediction with recurrent neural
  networks.
\newblock In \emph{Advances in Neural Information Processing Systems}, pages
  1171--1179, 2015.

\bibitem[Gao and Ganguli(2015)]{coupledneuron}
Peiran Gao and Surya Ganguli.
\newblock On simplicity and complexity in the brave new world of large-scale
  neuroscience.
\newblock \emph{Current opinion in neurobiology}, 32:\penalty0 148--155, 2015.

\bibitem[Gallego et~al.(2020)Gallego, Perich, Chowdhury, Solla, and
  Miller]{monkeys}
Juan~A Gallego, Matthew~G Perich, Raeed~H Chowdhury, Sara~A Solla, and Lee~E
  Miller.
\newblock Long-term stability of cortical population dynamics underlying
  consistent behavior.
\newblock \emph{Nature neuroscience}, 23\penalty0 (2):\penalty0 260--270, 2020.

\bibitem[Fey and Lenssen(2019)]{pyg}
Matthias Fey and Jan~E. Lenssen.
\newblock Fast graph representation learning with {PyTorch Geometric}.
\newblock In \emph{ICLR Workshop on Representation Learning on Graphs and
  Manifolds}, 2019.

\bibitem[Veli{\v{c}}kovi{\'c} et~al.(2017)Veli{\v{c}}kovi{\'c}, Cucurull,
  Casanova, Romero, Lio, and Bengio]{gat}
Petar Veli{\v{c}}kovi{\'c}, Guillem Cucurull, Arantxa Casanova, Adriana Romero,
  Pietro Lio, and Yoshua Bengio.
\newblock Graph attention networks.
\newblock \emph{arXiv preprint arXiv:1710.10903}, 2017.

\bibitem[Li et~al.(2015)Li, Tarlow, Brockschmidt, and Zemel]{gnngru}
Yujia Li, Daniel Tarlow, Marc Brockschmidt, and Richard Zemel.
\newblock Gated graph sequence neural networks.
\newblock \emph{arXiv preprint arXiv:1511.05493}, 2015.

\bibitem[Hu et~al.(2019)Hu, Liu, Gomes, Zitnik, Liang, Pande, and
  Leskovec]{transfergnn}
Weihua Hu, Bowen Liu, Joseph Gomes, Marinka Zitnik, Percy Liang, Vijay Pande,
  and Jure Leskovec.
\newblock Strategies for pre-training graph neural networks.
\newblock \emph{arXiv preprint arXiv:1905.12265}, 2019.

\bibitem[Horwitz(2003)]{funcvstruct}
Barry Horwitz.
\newblock The elusive concept of brain connectivity.
\newblock \emph{Neuroimage}, 19\penalty0 (2):\penalty0 466--470, 2003.

\end{thebibliography}

 
\section*{\Large{Supplementary Information}}

\section*{Model and Experiments}

\subsection*{Model Selection}
The two final models included in the main text were chosen for their performance and simplicity. Nevertheless, we experimented with numerous established models which were easily substituted for $f$. For GNNs, we primarily used the Pytorch Geometric library \citep{pyg}. Tested modules included the GIN-0/GIN-$\epsilon$ \citep{GIN}, Graph Sage \citep{graphsage}, GAT \citep{gat}, and Global Attention \citep{gnngru}. In particular, we expected the GIN to outperform the other modules because its expressiveness has been shown to aid transfer learning \citep{transfergnn}; however, because our edges are not explicitly known, we essentially applied the GIN on a fully connected graph. Under this formulation, the GIN-0 simply symmetrizes node features after a message passing step which is similar to the aggregation step of our MLP. We also found that the GIN-$\epsilon$ was prone to overfitting. Finally, we tested the GAT which is similar to our model when edges are dynamically inferred each timestep. As a result, we found that the GAT performs equally well on trajectory prediction but performs slightly worse on behavioral state classification.

\subsection*{Model Implementation}
The two-layer MLP corresponding to $g$ in the main text comprised of linear layers followed by ReLu activation functions. We also applied batch norm on the output of the two layers. The Node MLP in the main text refers to individual MLPs for each node. To construct RNN variants, we added an LTSM unit before the MLP.

We performed some minor hyperparameter optimization as our combinatorial cross-validation was computationally expensive. Overall, we found our models relatively robust to different hyperparameters. For trajectory prediction, we used hidden layers with 256 dimensions. On the other hand, for behavioral state classification, we used hidden layers with 16 dimensions. Furthermore, we determined that dynamic edges evaluation worked better for trajectory prediction; however, globally evaluated edges for each worm resulted in better performance for behavioral state classification. Finally, for trajectory prediction, we chose to optimize the mean square error (MSE). For behavioral state classification, we optimized the negative log likelihood (NLL). 

\subsection*{Experimental Procedures}
For the extended evaluation set, we chose data from the prelethargus phase, i.e. part of the stage of larval development associated with higher frequency pharyngeal pumping prior to a cessation during which the animal enters a brief lethargus, where 4 states were labeled: reverse, forward, dorsal turn, and ventral turn. For compatibility with the training dataset, we mapped reverse 1, reverse 2, and sustained reverse crawling to the reverse state. Similarly, we mapped forward crawling and forward slowing to forward. In addition to the 7 or 4 labeled states, there was another labeled state for unknown behavior or quiescence. This state comprised a very small portion of our data, and during training and evaluation, we ignore the result when the target is unknown.

For all experiments in the main text, we performed 10-fold cross validation on all possible permutations of worms in our training set (Kato dataset). For example, on our experiments trained on two worms, the possible permutations of worms are the following: $\displaystyle\{$(1, 2), (1, 3), (1, 4), (1, 5), (2, 3), (2, 4), (2, 5), (3, 4), (3, 5), (4, 5)$\}$. Experiments labeled with "Train on 2 worms" involved models trained separately on each of these permutations. Each permutation then involved 10-fold cross validation where the test set was left out when performing hyperparamter optimization. In particular, for our experiments on behavioral state classification, we used 1 fold as the test/"leave-out" set and 1 fold for the validation set which was used for optimization and as a metric for stopping training. On the other hand, our experiments on trajectory prediction was focused primarily on generalization performance instead of test set accuracy so we used 1 fold as the validation set and evaluated on all worms in the extended validation set (Nichols dataset). As a note, we also attempted experiments where data from the extended dataset was used as a validation set. Under this condition, we found that the MLP performed significantly better; however, we were concerned that the MLP was overfitting to the validation set so we chose not to included those results.

We performed our experiments on with an Intel i9 9900k CPU and Nvidia GeForce RTX 2080Ti graphics card. Since our models are relatively simple, we were able to train the model on data from one worm in one batch. Nevertheless, the number of worms and cross-validation procedure was very computationally expensive. As such, training and evaluating each model required roughly a week or two of continuous computation. For optimization, we used the Adams optimizer with a learning rate of $10^{-3}$. We decayed the learning rate with by a factor of 0.25 if the loss did not improve after 50 epochs. We then trained for 800 epoch and saved the model with the lowest validation loss. For scheduled sampling (used during trajectory prediction), we adopted a linear decay which terminated at 300 timesteps.

\subsection*{Additional Experiments}
We performed numerous experiments to verify our results and examine the performance of our model on diverse machine learning tasks. We did not perform rigorous cross validation for the following experiments.

\subsubsection*{Experiments without AVA}
Referees of \cite{brennan} were concerned with behavioral state classification where AVA neurons were included. In particular, these neurons were used by \cite{kato} to define behavioral state through trajectory clustering in latent space. Referees commented that classifying behavioral states with neurons used to define those states was akin to circular reasoning. We would like to note that \cite{kato} verified their assigned behavioral states through recorded videos, minimizing risks that assigned behavioural states differ from reality. Nevertheless, we followed \cite{brennan} and performed an experiment excluding AVA neurons in which we found no noticeable difference in model performance.

\subsubsection*{One-hot encoding of edges}
To enforce a sparsity on the edges, we experimented with one-hot encoding by adding a scaling factor within the softmax. We found that our GNN achieved similar test accuracies as in the main text. However, our GNN failed to generalize well to unseen worms. Following our discussion in the main text, we believe that one-hot encoding was detrimental to generalization because it effectively results in a permutation matrix which simply permutes node features. This is counter to previous studies where topological structures are related by more general linear transformations.

\begin{figure}[t!]
\renewcommand{\thefigure}{S1}
\begin{center}
\includegraphics[scale=.15]{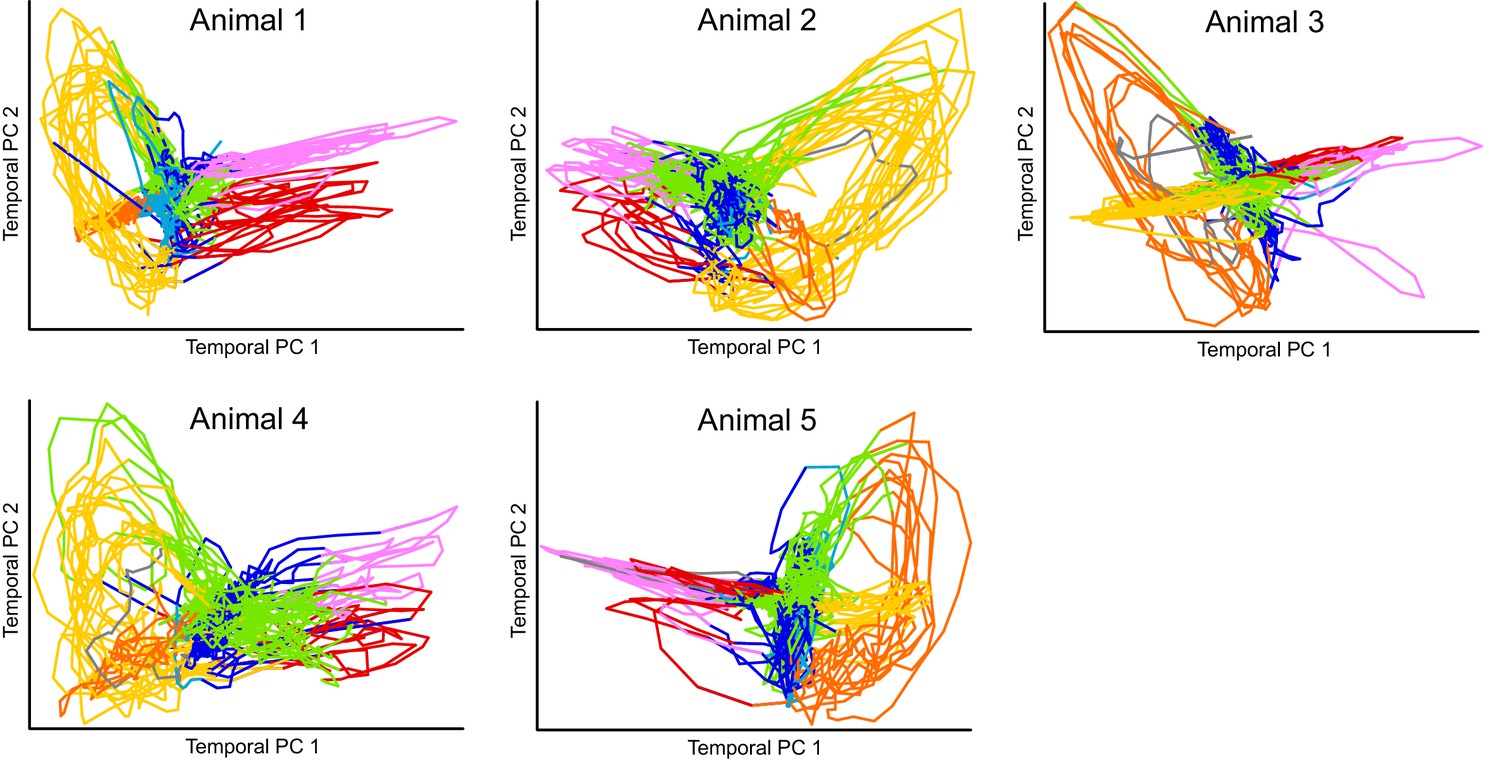}
\end{center}
\caption{Time derivatives of calcium traces projected onto each individual organism’s principal components. Distinct loops correspond to manifolds in latent space where colors correspond to behavior assigned in Kato et al. Reproduced with permission from \cite{brennan}.}
\label{fig:figureS1}
\end{figure}

\subsubsection*{Comparison of inferred edges to known connectome}
\label{sec:funcvstruc}
Inferring the connectivity between neurons in neural systems remains a key challenge in neuroscience. Because C. Elegans is among few organisms whose connectome mostly or completely known, we decided to compare the inferred edges of our model to the connectome of C. Elegans. Ultimately, we found no similarities between our inferred edges and the connectome.

In neuroscience, two types of connectivity are defined: structural and functional/effective. Structural connectivity refers to physical connections between neurons, whereas functional connectivity implies statistical correlations between neurons and effective connectivity validated causal connections between neurons \citep{funcvstruct}. The development of methods for determining functional and, in particular, effective connectivity remains an open challenge and a highly active area of research . Nonetheless, in the context of C. elegans, each worm generally has the same structural connectivity; however, differences in neural activity implies a different functional connectivity exists for unique individuals. Since the connectome relates to the structural connectivity, we believe that our inferred edges are a poor proxy for the connectome. On a more abstract level, our graph neural network works with a subset of neurons such that a inferred edge may not correspond to a direct correlation, but may rather represent higher order correlations with unseen neurons.

\end{document}